\documentclass[sigconf, review=false, screen]{acmart}
\usepackage{booktabs}            
\usepackage{multirow}            
\usepackage{tabularx}
\usepackage{tabularx} 
\usepackage{soul}
\AtBeginDocument{%
  }

\setcopyright{acmlicensed}
\copyrightyear{2018}
\acmYear{2018}
\acmDOI{XXXXXXX.XXXXXXX}
\acmConference[Conference acronym 'XX]{Make sure to enter the correct
  conference title from your rights confirmation email}{June 03--05,
  2018}{Woodstock, NY}
\acmISBN{978-1-4503-XXXX-X/2018/06}

\begin{document}

\title[InforID]{Adaptive Semantic Capacity Allocation for Parallel Generative Recommendation}

\author{Chenxi Li}
\email{lichenxi24@mails.ucas.ac.cn}
\affiliation{%
  \institution{University of the Chinese Academy of Sciences}
  \city{Beijing}
  \country{China}
}
\affiliation{%
  \institution{Institute of Automation, Chinese Academy of Sciences}
  \city{Beijing}
  \country{China}
}

\author{Yuchen Lu}
\email{luyuchen2021@ia.ac.cn}
\affiliation{%
  \institution{Institute of Automation, Chinese Academy of Sciences}
  \city{Beijing}
  \country{China}
}
\affiliation{%
  \institution{University of the Chinese Academy of Sciences}
  \city{Beijing}
  \country{China}
}

\author{Xu Yang}
\email{xu.yang@ia.ac.cn}
\affiliation{%
  \institution{Institute of Automation, Chinese Academy of Sciences}
  \city{Beijing}
  \country{China}
}
\affiliation{%
  \institution{University of the Chinese Academy of Sciences}
  \city{Beijing}
  \country{China}
}

\thanks{* Corresponding author: Xu Yang (xu.yang@ia.ac.cn).}

\renewcommand{\shortauthors}{Chenxi Li et al.}

\begin{abstract}
Autoregressive semantic ID recommenders are constrained by expensive beam-search decoding, which limits the practical length of item identifiers. Parallel generation methods alleviate this bottleneck by predicting all semantic ID tokens simultaneously, enabling longer IDs. However, existing semantic ID methods still rely on manually predefined and homogeneous ID structures, where both the number of semantic slots and the codebook size of each slot are treated as fixed hyperparameters. This ignores the heterogeneous capacity demands of different semantic subspaces and may allocate prediction capacity to slots with limited utility. We show that uniformly expanding semantic slots can provide limited gains, indicating redundant capacity in homogeneous semantic IDs. We propose InforID, a lightweight adaptive semantic target construction framework for parallel generative recommendation. InforID allocates a fixed capacity budget across candidate semantic slots, thereby jointly determining the effective ID length and slot-specific codebook sizes. Experiments demonstrate improved recommendation accuracy under comparable capacity budgets while preserving one-step parallel prediction. Code is available at \url{https://anonymous.4open.science/r/inforID-F582}.
\end{abstract}





\maketitle

\section{Introduction}
Semantic ID-based generative recommendation represents each item with a discrete semantic identifier and predicts target items in the semantic ID space~\cite{tiger, p5, hstu, rpg}. Unlike atomic item IDs, semantic IDs decompose an item into multiple discrete semantic tokens, enabling recommendation models to capture sub-item semantic structure~\cite{vqrec,tiger}. However, autoregressive semantic ID recommenders decode identifiers token by token and typically rely on beam search, which makes inference expensive and restricts the practical length of item identifiers~\cite{seq2seq,beam,tiger}. Recent parallel generation methods alleviate this bottleneck by predicting all semantic ID tokens simultaneously, making long semantic identifiers feasible~\cite{rpg, gptrec}.

Despite this progress, the structure of semantic IDs remains largely manually predefined. Existing semantic ID methods usually determine the number of semantic slots and the codebook size of each slot through hyperparameter selection~\cite{vqrec,tiger,rpg}. In particular, parallel semantic ID methods commonly use homogeneous codebooks, assigning the same number of codewords to every slot~\cite{rpg}. This design implicitly assumes that all semantic subspaces require equal representational capacity. We argue that this homogeneous capacity assumption is often suboptimal.

\begin{table}[t]
\setlength{\tabcolsep}{3pt} 
\centering
\caption{Effect of uniform slot expansion on Sports. Each slot has 32 dimensions and 256 codewords.}
\label{tab:slot_expansion}
\scriptsize
\begin{tabular}{c c c c c c}
\toprule
\#Slots & Dim. & R@5 & N@5 & R@10 & N@10 \\
\midrule
13 & 416 & 0.0285 & 0.0201 & 0.0430 & 0.0248 \\
14 & 448 & \textbf{0.0307} & \textbf{0.0211} & \textbf{0.0464} & \textbf{0.0261} \\
15 & 480 & 0.0301 & \textbf{0.0211} & 0.0452 & 0.0259 \\
16 & 512 & 0.0299 & 0.0205 & 0.0453 & 0.0254 \\
17 & 544 & 0.0290 & 0.0200 & 0.0442 & 0.0249 \\
18 & 576 & 0.0300 & 0.0208 & 0.0458 & 0.0252 \\
\bottomrule
\end{tabular}
\addvspace{-12pt}
\end{table}

Simply increasing the number of uniformly allocated semantic slots does not necessarily improve recommendation. Although OPQ rotates item representations to make subspaces more suitable for product quantization~\cite{opq,pq}, the number of slots and the codebook size of each slot remain manually fixed. In our preliminary study, reducing the retained projected dimensionality before uniform partitioning causes little performance degradation and sometimes even improves accuracy. This suggests that simply adding more uniformly parameterized slots does not guarantee more useful prediction targets, and that part of the codebook capacity may be spent on slots that contribute little to recommendation.

Motivated by these findings, we revisit semantic ID construction as a target-space capacity allocation problem for parallel generative recommendation. Our goal is not to propose a new quantization algorithm for embedding compression, but to construct a discrete semantic target space that better matches the heterogeneous capacity demands of candidate slots. We propose InforID, a lightweight and pluggable construction framework that adaptively allocates capacity over candidate semantic slots. The resulting allocation determines not only slot-specific codebook sizes, but also the effective ID length: slots assigned zero bits are removed because their single-codeword codebooks cannot distinguish items. The retained slots are then predicted in parallel with slot-specific heads, preserving the efficiency advantage of parallel generation~\cite{rpg}.

Our contributions are threefold. \textbf{First}, we identify the manually predefined homogeneous ID structure as an overlooked limitation of semantic ID recommendation. \textbf{Second}, we propose InforID, a lightweight adaptive target construction framework for parallel generative recommendation that determines slot-specific codebook sizes and a data-dependent effective ID length. \textbf{Third}, experiments show that InforID improves recommendation accuracy under comparable capacity budgets while preserving parallel prediction.
\section{Motivation}

\begin{figure}[t]
    \centering
    \Description{Two plots showing energy distribution. Plot (a) shows a sharp decay in eigenvalues of principal components across four datasets. Plot (b) shows that cumulative variance reaches 80 percent with only a small fraction of dimensions.}
    \includegraphics[width=0.9\linewidth, height=0.3\linewidth]{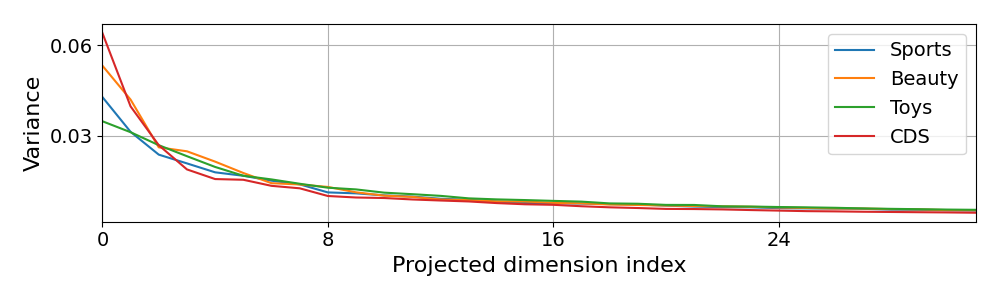}
    \caption{Information distribution across projected subspaces. Skewed energy indicates heterogeneous capacity demands.}
    \label{fig:energy_analysis}
\addvspace{-15pt}
\end{figure}

We first examine whether uniformly increasing semantic ID length consistently
improves recommendation. In an OPQ-based parallel semantic ID construction
pipeline~\cite{opq,rpg}, item representations are projected and partitioned into $m$ equal-size
subspaces, where each slot has 32 dimensions and uses a codebook size of 256.
As shown in Table~\ref{tab:slot_expansion}, increasing $m$ from 13 to 18 on the
Sports dataset does not yield monotonic gains: performance peaks at $m=14$ and
then fluctuates or degrades. This suggests that simply adding homogeneous slots
does not necessarily produce more useful prediction targets.

To explain why homogeneous capacity may be inefficient, we analyze the
information distribution before ID construction. After projection, we measure
the energy of each candidate subspace by
\[
s_j = \sum_{d \in G_j} \lambda_d,
\]
where $G_j$ is the $j$-th subspace and $\lambda_d$ is the variance of the $d$-th
projected dimension. Figure~\ref{fig:energy_analysis} shows a highly skewed
energy distribution, indicating that different semantic slots have heterogeneous
capacity demands. This motivates adaptive semantic target construction instead of manually fixed homogeneous ID structures.

\begin{figure*}[t] 
    \centering
    \includegraphics[width=0.7\linewidth, height=0.35\linewidth]{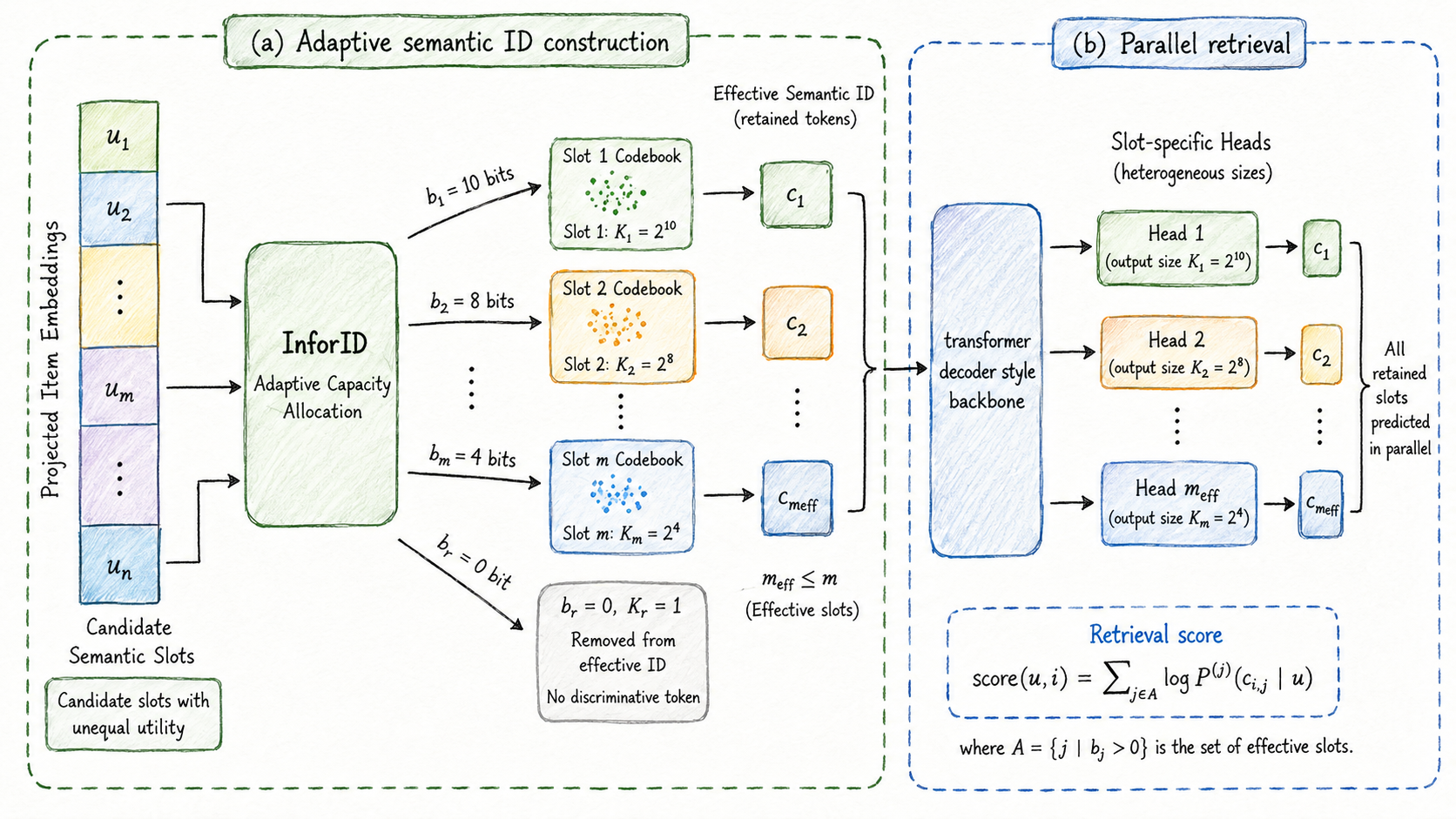} 
    \Description{A diagram showing InforID. Projected item embeddings are divided into candidate semantic slots. Adaptive capacity allocation assigns different bit-widths, removes zero-bit slots, and forms heterogeneous semantic IDs, which are predicted in parallel by slot-specific heads.}
    \caption{Overview of InforID. Adaptive capacity allocation determines heterogeneous semantic IDs, removes zero-bit slots.}
    \label{fig:architecture}
\end{figure*}

\section{Method}

InforID is designed as an adaptive semantic target construction framework for parallel generative recommendation. Given candidate semantic slots, it allocates a fixed capacity budget to determine which slots are retained and how large their slot-specific vocabularies are. The resulting semantic IDs define the prediction targets, the output dimensions of parallel heads, and the item scoring space.

\subsection{Candidate Semantic Slots}

Let $\mathbf{x}_i \in \mathbb{R}^{D}$ denote the continuous semantic representation of item $i$, extracted from item content such as text or multimodal features~\cite{vqrec,tiger,rpg, vqvae}. Directly partitioning the raw embedding dimensions may lead to highly correlated and unstable subspaces. Therefore, we first transform item representations into a projected space to obtain a more suitable basis for semantic slot construction~\cite{pq,opq,rpg}. The projected representation is then divided into $M$ equal-size subspaces:
\[
\mathbf{z}_i = [\mathbf{z}_i^{(1)}, \mathbf{z}_i^{(2)}, \ldots, \mathbf{z}_i^{(M)}],
\]
where $\mathbf{z}_i^{(j)}$ denotes the sub-vector of item $i$ in the $j$-th subspace. Each subspace corresponds to a candidate semantic slot.

Different from homogeneous semantic ID construction, InforID does not force all candidate slots to appear in the final identifier with the same capacity. Instead, these slots serve as candidates for subsequent capacity allocation. A slot will be retained in the effective semantic ID only if it receives positive capacity. Therefore, the final ID structure is not manually fixed in advance, but determined by the adaptive allocation process described next.

\subsection{Adaptive Capacity Allocation}

Given the candidate semantic slots, InforID assigns each slot a non-negative bit-width $b_j$, which determines its codebook size $K_j=2^{b_j}$. Instead of using the same bit-width for all slots, InforID allocates a fixed total budget $B$ across candidate slots:
\[
\sum_{j=1}^{M} b_j = B, \quad b_j \in \mathbb{Z}_{\ge 0}.
\]

For the $j$-th candidate slot, let $\mathcal{C}_j(b_j)$ denote the codebook with $2^{b_j}$ codewords obtained by k-means clustering on the corresponding subspace~\cite{kmeans}. We define its subspace reconstruction loss as
\[
\mathcal{L}_j(b_j)
=
\frac{1}{|\mathcal{I}|}
\sum_{i\in\mathcal{I}}
\min_{\mathbf{e}\in\mathcal{C}_j(b_j)}
\left\|
\mathbf{z}_i^{(j)}-\mathbf{e}
\right\|_2^2 .
\]

Here, reconstruction loss serves only as a lightweight proxy for preserving item-level semantic distinctions during target construction; the final criterion remains downstream retrieval performance.

The capacity allocation objective is
\[
\min_{\{b_j\}_{j=1}^{M}}
\sum_{j=1}^{M} \mathcal{L}_j(b_j),
\quad
\mathrm{s.t.}\quad
\sum_{j=1}^{M} b_j = B,\;
b_j \in \mathbb{Z}_{\ge 0}.
\]

Since exhaustive search over all bit allocations is impractical, we use a greedy allocation procedure. Starting from $b_j=0$ for all slots, InforID assigns bits one at a time. At each iteration, it tentatively adds one bit to each candidate slot and computes the marginal reduction in reconstruction loss:
\[
\Delta_j
=
\mathcal{L}_j(b_j)
-
\mathcal{L}_j(b_j+1).
\]
The bit is assigned to the slot with the largest reduction:
\[
j^*=\arg\max_j \Delta_j,
\quad
b_{j^*}\leftarrow b_{j^*}+1.
\]
This process repeats until the total budget $B$ is exhausted. The final allocation yields heterogeneous codebook sizes across slots. Importantly, it also induces slot selection: slots assigned zero bits are removed from the final semantic ID, making the effective ID length data-dependent.

\subsection{Parallel Retrieval with Heterogeneous Semantic IDs}

After allocation, InforID retains only the slots assigned positive bit-widths:
\[
\mathcal{A}=\{j\mid b_j>0\}, \quad
m_{\mathrm{eff}}=|\mathcal{A}|.
\]
For each retained slot $j\in\mathcal{A}$, its codebook $\mathcal{C}_j=\{\mathbf{e}_{j,c}\}_{c=1}^{K_j}$ has size $K_j=2^{b_j}$. The semantic token of item $i$ at slot $j$ is the index of the nearest codeword:
\[
c_{i,j}
=
\arg\min_{c\in\{1,\ldots,K_j\}}
\left\|
\mathbf{z}_i^{(j)}-\mathbf{e}_{j,c}
\right\|_2^2 .
\]
The final semantic ID of item $i$ is
\[
\mathbf{c}_i=(c_{i,j})_{j\in\mathcal{A}}.
\]
If $b_j=0$, then $K_j=1$, so all items share the same codeword in this slot. Such a slot provides no discriminative information and is excluded from the effective semantic ID, prediction, and scoring.

Given a user history, a transformer decoder-style backbone encodes it into $\mathbf{h}_u$~\cite{transformer}. Following the parallel prediction paradigm, we factorize the probability of the target semantic ID over retained slots~\cite{rpg,mtp}:
\[
P(\mathbf{c}_{i^+}\mid u)
=
\prod_{j\in\mathcal{A}}
P^{(j)}(c_{i^+,j}\mid u).
\]
Each retained slot uses a slot-specific head to output~\cite{medusa,simclr,rpg}
\[
P^{(j)}(c\mid u), \quad c\in\{1,\ldots,K_j\}.
\]
Thus, the output dimension of each head is determined by the allocated codebook size $K_j$.

Given the ground-truth next item $i^+$, the multi-token prediction loss is
\[
\mathcal{L}_{\mathrm{MTP}}
=
-\sum_{j\in\mathcal{A}}
\log P^{(j)}(c_{i^+,j}\mid u).
\]
At inference time, candidate item $i$ is scored using the same additive log-probability form~\cite{rpg}:
\[
\mathrm{score}(u,i)
=
\sum_{j\in\mathcal{A}}
\log P^{(j)}(c_{i,j}\mid u).
\]
All ID construction is performed offline. Online inference follows the same one-step parallel prediction paradigm~\cite{rpg,nat,mask}, so InforID changes the effective ID length, slot-specific vocabularies, and scoring space without introducing sequential decoding overhead.

\begin{table*}[t]
\renewcommand{\arraystretch}{1.3}
  \caption{Performance comparison between InforID and various baselines across four datasets. The best results are highlighted in bold, and the second-best results are underlined.}
  \label{tab:multi_row_example}
  \centering
  \footnotesize 
  \setlength{\tabcolsep}{0pt} 
  
  \begin{tabular*}{\textwidth}{@{\extracolsep{\fill}} l rrrr rrrr rrrr rrrr}
    \toprule
    \multirow{2}{*}{\textbf{Method}} & \multicolumn{4}{c}{Sports} & \multicolumn{4}{c}{Beauty} & \multicolumn{4}{c}{Toys} & \multicolumn{4}{c}{CDs} \\
    \cmidrule(lr){2-5} \cmidrule(lr){6-9} \cmidrule(lr){10-13} \cmidrule(lr){14-17}
    & R@5 & N@5 & R@10 & N@10 & R@5 & N@5 & R@10 & N@10 & R@5 & N@5 & R@10 & N@10 & R@5 & N@5 & R@10 & N@10 \\
    \midrule \midrule
    SASRec & 0.0233 & 0.0154 & 0.0350 & 0.0192 & 0.0387 & 0.0249 & 0.0605 & 0.0318 & 0.0463 & 0.0306 & 0.0675 & 0.0374 & 0.0351 & 0.0177 & 0.0619 & 0.0263 \\
    RecJPQ & 0.0141 & 0.0076 & 0.0220 & 0.0102 & 0.0311 & 0.0167 & 0.0482 & 0.0222 & 0.0331 & 0.0182 & 0.0484 & 0.0231 & 0.0075 & 0.0046 & 0.0138 & 0.0066 \\
    VQ-Rec & 0.0208 & 0.0144 & 0.0300 & 0.0173 & 0.0457 & 0.0317 & 0.0664 & 0.0383 & 0.0497 & 0.0346 & 0.0737 & 0.0230 & 0.0352 & 0.0238 & 0.0520 & 0.0292 \\
    TIGER & 0.0264 & 0.0181 & 0.0400 & 0.0225 & 0.0454 & 0.0321 & 0.0648 & 0.0384 & 0.0521 & 0.0371 & 0.0712 & 0.0432 & 0.0492 & 0.0329 & \ul{0.0748} & 0.0411 \\
    RPG & \ul{0.0314} & \ul{0.0216} & \ul{0.0463} & \ul{0.0263} & \ul{0.0550} & \ul{0.0381} & \ul{0.0809} & \ul{0.0464} & \ul{0.0592} & \ul{0.0401} & \ul{0.0869} & \ul{0.0490} & \ul{0.0498} & \ul{0.0338} & 0.0735 & \ul{0.0415} \\
    \textbf{InforID} & \textbf{0.0329} & \textbf{0.0229} & \textbf{0.0491} & \textbf{0.0279} & \textbf{0.0562} & \textbf{0.0388} & \textbf{0.0815} & \textbf{0.0469} & \textbf{0.0613} & \textbf{0.0421} & \textbf{0.0877} & \textbf{0.0506} & \textbf{0.0521} & \textbf{0.0351} & \textbf{0.0764} & \textbf{0.0431} \\
    \bottomrule
  \end{tabular*}
\end{table*}

\section{EXPERIMENTS}
\subsection{Experimental Setup}

We evaluate InforID on four public Amazon review benchmarks: Sports, Beauty,
Toys, and CDs~\cite{amazon}, using Recall and NDCG at cutoffs 5 and 10 for
next-item retrieval evaluation. We compare with SASRec~\cite{sasrec},
VQ-Rec~\cite{vqrec}, RecJPQ~\cite{recjpq}, TIGER~\cite{tiger}, and
RPG~\cite{rpg}, covering item-ID based recommendation, quantized item
representation, autoregressive semantic ID generation, and parallel semantic
ID generation. All reported results are averaged over three runs with different
random seeds.

For item representation, TIGER uses sentence-t5-base~\cite{sentence_t5,tiger}
following its original setting, whereas RPG and InforID use
text-embedding-3-large~\cite{textemb}. To ensure that the comparison is not
tied to a specific embedding source, we further evaluate TIGER, RPG, and
InforID with both embeddings and observe similar relative trends. To isolate
the effect of semantic ID construction, RPG and InforID share the same item
embeddings and parallel prediction backbone. InforID only replaces RPG's
manually fixed homogeneous ID construction with adaptive semantic capacity
allocation, and ID construction variants are compared under comparable capacity
budgets unless otherwise specified.
\begin{table}[t]
\centering
\caption{Ablation study on semantic ID construction strategies under the same bit budget. 
NDCG@10 and relative reconstruction loss are reported for each dataset.}
\label{tab:ablation_id}
\scriptsize
\setlength{\tabcolsep}{3pt}
\begin{tabular}{lcccccccc}
\toprule
\multirow{2}{*}{\textbf{Variant}} 
& \multicolumn{2}{c}{\textbf{Sports}} 
& \multicolumn{2}{c}{\textbf{Beauty}} 
& \multicolumn{2}{c}{\textbf{Toys}} 
& \multicolumn{2}{c}{\textbf{CDs}} \\
\cmidrule(lr){2-3}
\cmidrule(lr){4-5}
\cmidrule(lr){6-7}
\cmidrule(lr){8-9}
& \textbf{Loss $\downarrow$} & \textbf{N@10 $\uparrow$} 
& \textbf{Loss $\downarrow$} & \textbf{N@10 $\uparrow$} 
& \textbf{Loss $\downarrow$} & \textbf{N@10 $\uparrow$} 
& \textbf{Loss $\downarrow$} & \textbf{N@10 $\uparrow$} \\
\midrule
PQ     
& 1.000 & 0.0239 
& 1.000 & 0.0412 
& 1.000 & 0.0425 
& 1.000 & 0.0374 \\
OPQ    
& 0.775 & 0.0255 
& 0.657 & 0.0464 
& 0.771 & 0.0488 
& 0.738 & 0.0386 \\
\midrule
\textbf{InforID} 
& \textbf{0.583} & \textbf{0.0279} 
& \textbf{0.475} & \textbf{0.0469} 
& \textbf{0.570} & \textbf{0.0506} 
& \textbf{0.583} & \textbf{0.0431} \\
\bottomrule
\end{tabular}
\addvspace{-13pt}
\end{table}

\subsection{Overall Performance}
Table~\ref{tab:multi_row_example} reports the overall performance. InforID
achieves the best or comparable results across the four benchmarks. Compared
with SASRec~\cite{sasrec} and quantized representation baselines~\cite{vqrec,recjpq}, semantic ID-based generative
methods~\cite{tiger,rpg} show stronger performance, indicating the effectiveness of
content-derived discrete identifiers.

RPG is the most relevant baseline because it uses the same parallel prediction
paradigm as InforID. Under the same item embeddings and backbone, InforID achieves better or comparable performance than RPG across all four datasets. This suggests that the gain mainly comes
from replacing the manually fixed homogeneous ID structure with adaptive semantic
target construction.

\subsection{Capacity Allocation Analysis}

We further analyze whether the improvement of InforID comes from adaptive
capacity allocation. To this end, we compare different semantic ID construction
strategies under the same total capacity budget, including uniform product
quantization (PQ), optimized product quantization (OPQ), and InforID. All
methods use the same item embeddings and parallel prediction backbone, and
differ only in how semantic IDs are constructed.

Table~\ref{tab:ablation_id} reports both NDCG@10 and relative reconstruction
loss for each dataset. The relative reconstruction loss is computed over all candidate subspaces and
normalized by the PQ loss on each dataset under the same bit budget. PQ and OPQ construct content-based semantic IDs,
but they still rely on homogeneous slot capacities, assigning the same
vocabulary size to each semantic slot. In contrast, InforID adaptively reshapes
the semantic target space by assigning different vocabulary sizes to retained
slots and removing zero-bit slots from prediction and scoring.

InforID achieves consistently strong reconstruction quality and downstream
NDCG@10. This supports using marginal reconstruction gain as a lightweight
allocation signal: it helps preserve item-level semantic distinctions during
target construction, while the final criterion remains downstream retrieval
performance. 

These controlled comparisons indicate that the benefit of InforID does not
come from a different recommender backbone or item embedding source, but from
constructing a more suitable heterogeneous semantic target space under the same
prediction paradigm and capacity budget.

We additionally observe lower ID collision rates than homogeneous OPQ under
the same bit budget, suggesting improved item discriminability.

\subsection{Budget Sensitivity}
\begin{figure}[t]
    \centering
    \Description{Line charts showing NDCG at 10 performance across four datasets as bit count increases from 64 to 352. All curves show a sharp initial rise followed by a plateau.}
    \includegraphics[width=1\linewidth]{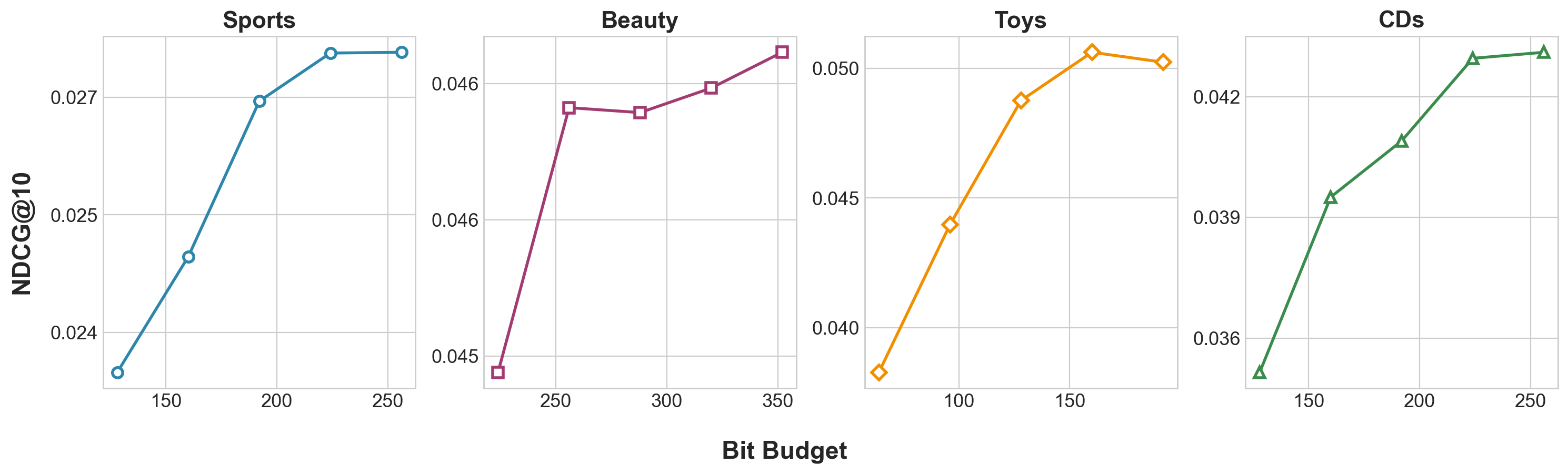}
    \caption{Sensitivity analysis of total bit budget $B$ on four datasets. The NDCG@10 performance is reported as $B$ varies from 64 to 352.}
    \label{fig:bit_sensitivity}
\end{figure}

We further vary the total capacity budget $B$ to examine the robustness of InforID. As shown in Figure~\ref{fig:bit_sensitivity}, performance improves when the budget increases but gradually saturates, indicating diminishing returns of semantic capacity. 

\section{Conclusion}
We revisit semantic ID construction as adaptive semantic target construction for parallel generative recommendation. InforID allocates capacity across candidate semantic slots, jointly determining the effective ID length and slot-specific codebook sizes while preserving one-step parallel prediction. Experiments show that semantic ID structures should be data-dependent rather than manually fixed and homogeneous.

\section*{GenAI Usage Disclosure}

The authors did not use generative AI tools for code implementation, data processing, model training, hyperparameter tuning, evaluation, or conducting experiments. During manuscript preparation, generative AI tools were used only for minor grammar correction, sentence-level phrasing refinement, and style transfer of the architecture figure for visual presentation purposes. These tools were not used to generate scientific claims, experimental results, or technical conclusions. All research ideas, technical contributions, methodological design, and experimental analysis were developed, validated, and written by the authors.

\appendix
\bibliographystyle{ACM-Reference-Format}
\bibliography{sample-base}

@inproceedings{vqrec,
  title={Learning vector-quantized item representation for transferable sequential recommenders},
  author={Hou, Yupeng and He, Zhankui and McAuley, Julian and Zhao, Wayne Xin},
  booktitle={Proceedings of the ACM Web Conference 2023},
  pages={1162--1171},
  year={2023}
}

@article{tiger,
  title={Recommender systems with generative retrieval},
  author={Rajput, Shashank and Mehta, Nikhil and Singh, Anima and Hulikal Keshavan, Raghunandan and Vu, Trung and Heldt, Lukasz and Hong, Lichan and Tay, Yi and Tran, Vinh and Samost, Jonah and others},
  journal={Advances in Neural Information Processing Systems},
  volume={36},
  pages={10299--10315},
  year={2023}
}

@inproceedings{rpg,
  title={Generating long semantic ids in parallel for recommendation},
  author={Hou, Yupeng and Li, Jiacheng and Shin, Ashley and Jeon, Jinsung and Santhanam, Abhishek and Shao, Wei and Hassani, Kaveh and Yao, Ning and McAuley, Julian},
  booktitle={Proceedings of the 31st ACM SIGKDD Conference on Knowledge Discovery and Data Mining V. 2},
  pages={956--966},
  year={2025}
}

@article{pq,
  title={Product quantization for nearest neighbor search},
  author={Jegou, Herve and Douze, Matthijs and Schmid, Cordelia},
  journal={IEEE transactions on pattern analysis and machine intelligence},
  volume={33},
  number={1},
  pages={117--128},
  year={2010},
  publisher={IEEE}
}

@article{opq,
  title={Optimized product quantization},
  author={Ge, Tiezheng and He, Kaiming and Ke, Qifa and Sun, Jian},
  journal={IEEE transactions on pattern analysis and machine intelligence},
  volume={36},
  number={4},
  pages={744--755},
  year={2013},
  publisher={IEEE}
}

@inproceedings{kmeans,
  title={Some methods of classification and analysis of multivariate observations},
  author={McQueen, James B},
  booktitle={Proc. of 5th Berkeley Symposium on Math. Stat. and Prob.},
  pages={281--297},
  year={1967}
}

@article{mtp,
  title={Better \& faster large language models via multi-token prediction},
  author={Gloeckle, Fabian and Idrissi, Badr Youbi and Rozi{\`e}re, Baptiste and Lopez-Paz, David and Synnaeve, Gabriel},
  journal={arXiv preprint arXiv:2404.19737},
  year={2024}
}

@inproceedings{amazon,
  title={Ups and downs: Modeling the visual evolution of fashion trends with one-class collaborative filtering},
  author={He, Ruining and McAuley, Julian},
  booktitle={proceedings of the 25th international conference on world wide web},
  pages={507--517},
  year={2016}
}

@inproceedings{sasrec,
  title={Self-attentive sequential recommendation},
  author={Kang, Wang-Cheng and McAuley, Julian},
  booktitle={2018 IEEE international conference on data mining (ICDM)},
  pages={197--206},
  year={2018},
  organization={IEEE}
}

@inproceedings{recjpq,
  title={RecJPQ: training large-catalogue sequential recommenders},
  author={Petrov, Aleksandr V and Macdonald, Craig},
  booktitle={Proceedings of the 17th ACM International Conference on Web Search and Data Mining},
  pages={538--547},
  year={2024}
}

@inproceedings{sentence_t5,
  title={Sentence-t5: Scalable sentence encoders from pre-trained text-to-text models},
  author={Ni, Jianmo and Abrego, Gustavo Hernandez and Constant, Noah and Ma, Ji and Hall, Keith and Cer, Daniel and Yang, Yinfei},
  booktitle={Findings of the association for computational linguistics: ACL 2022},
  pages={1864--1874},
  year={2022}
}

@article{transformer,
  title={Attention is all you need},
  author={Vaswani, Ashish and Shazeer, Noam and Parmar, Niki and Uszkoreit, Jakob and Jones, Llion and Gomez, Aidan N and Kaiser, {\L}ukasz and Polosukhin, Illia},
  journal={Advances in neural information processing systems},
  volume={30},
  year={2017}
}

@inproceedings{p5,
  title={Recommendation as language processing (rlp): A unified pretrain, personalized prompt \& predict paradigm (p5)},
  author={Geng, Shijie and Liu, Shuchang and Fu, Zuohui and Ge, Yingqiang and Zhang, Yongfeng},
  booktitle={Proceedings of the 16th ACM conference on recommender systems},
  pages={299--315},
  year={2022}
}

@article{hstu,
  title={Actions speak louder than words: Trillion-parameter sequential transducers for generative recommendations},
  author={Zhai, Jiaqi and Liao, Lucy and Liu, Xing and Wang, Yueming and Li, Rui and Cao, Xuan and Gao, Leon and Gong, Zhaojie and Gu, Fangda and He, Michael and others},
  journal={arXiv preprint arXiv:2402.17152},
  year={2024}
}

@article{vqvae,
  title={Neural discrete representation learning},
  author={Van Den Oord, Aaron and Vinyals, Oriol and others},
  journal={Advances in neural information processing systems},
  volume={30},
  year={2017}
}

@article{medusa,
  title={Medusa: Simple llm inference acceleration framework with multiple decoding heads},
  author={Cai, Tianle and Li, Yuhong and Geng, Zhengyang and Peng, Hongwu and Lee, Jason D and Chen, Deming and Dao, Tri},
  journal={arXiv preprint arXiv:2401.10774},
  year={2024}
}

@inproceedings{simclr,
  title={A simple framework for contrastive learning of visual representations},
  author={Chen, Ting and Kornblith, Simon and Norouzi, Mohammad and Hinton, Geoffrey},
  booktitle={International conference on machine learning},
  pages={1597--1607},
  year={2020},
  organization={PmLR}
}

@inproceedings{textemb,
  title={Improving text embeddings with large language models},
  author={Wang, Liang and Yang, Nan and Huang, Xiaolong and Yang, Linjun and Majumder, Rangan and Wei, Furu},
  booktitle={Proceedings of the 62nd Annual Meeting of the Association for Computational Linguistics (Volume 1: Long Papers)},
  pages={11897--11916},
  year={2024}
}

@inproceedings{mask,
  title={Mask-predict: Parallel decoding of conditional masked language models},
  author={Ghazvininejad, Marjan and Levy, Omer and Liu, Yinhan and Zettlemoyer, Luke},
  booktitle={Proceedings of the 2019 conference on empirical methods in natural language processing and the 9th international joint conference on natural language processing (EMNLP-IJCNLP)},
  pages={6112--6121},
  year={2019}
}

@article{nat,
  title={Non-autoregressive neural machine translation},
  author={Gu, Jiatao and Bradbury, James and Xiong, Caiming and Li, Victor OK and Socher, Richard},
  journal={arXiv preprint arXiv:1711.02281},
  year={2017}
}

@article{gptrec,
  title={Generative sequential recommendation with gptrec},
  author={Petrov, Aleksandr V and Macdonald, Craig},
  journal={arXiv preprint arXiv:2306.11114},
  year={2023}
}

@article{seq2seq,
  title={Sequence to sequence learning with neural networks},
  author={Sutskever, Ilya and Vinyals, Oriol and Le, Quoc V},
  journal={Advances in neural information processing systems},
  volume={27},
  year={2014}
}

@article{beam,
  title={Google's neural machine translation system: Bridging the gap between human and machine translation},
  author={Wu, Yonghui and Schuster, Mike and Chen, Zhifeng and Le, Quoc V and Norouzi, Mohammad and Macherey, Wolfgang and Krikun, Maxim and Cao, Yuan and Gao, Qin and Macherey, Klaus and others},
  journal={arXiv preprint arXiv:1609.08144},
  year={2016}
}


\end{document}